\documentclass{article}

\PassOptionsToPackage{numbers, compress}{natbib}
\usepackage[final]{neurips_2021}

\usepackage[utf8]{inputenc} % allow utf-8 input
\usepackage[T1]{fontenc}    % use 8-bit T1 fonts
\usepackage{hyperref}       % hyperlinks
\usepackage{url}            % simple URL typesetting
\usepackage{booktabs}       % professional-quality tables
\usepackage{amsfonts}       % blackboard math symbols
\usepackage{nicefrac}       % compact symbols for 1/2, etc.
\usepackage{microtype}      % microtypography
\usepackage[dvipsnames]{xcolor}  % colors
\usepackage{graphicx,epsfig} % graphics
\usepackage{multirow}

\usepackage{colortbl}

\usepackage{amsthm}
\newtheorem{definition}{Definition}
\newtheorem{proposition}{Proposition}

\usepackage{amsmath}

\usepackage{pifont}% http://ctan.org/pkg/pifont
\newcommand{\cmark}{\ding{51}}%
\newcommand{\xmark}{\ding{55}}%

\usepackage{pst-solides3d}

\title{Latent Space Refinement for Deep Generative Models}

\author{%
	Ramon Winterhalder,$^{a,b,c}$ Marco Bellagente,$^{a}$, and Benjamin Nachman$^{d,e}$\\
	\phantom{ }\hspace{-0.12in}$^a$Institute for Theoretical Physics, Heidelberg University, Germany \\
	\phantom{ }\hspace{-0.12in}$^b$Institute for Theoretical Physics, Karlsruhe Institute for Technology, Germany \\
	\phantom{ }\hspace{-0.12in}$^c$Centre for Cosmology, Particle Physics and Phenomenology (CP3), \\Universit\'e catholique de Louvain, Belgium\\
	\phantom{ }\hspace{-0.12in}$^d$Physics Division, Lawrence Berkeley National Laboratory, Berkeley, CA 94720, USA \\
	\phantom{ }\hspace{-0.12in}$^e$Berkeley Institute for Data Science, University of California, Berkeley, CA 94720, USA \\
	\\
	\texttt{ramon.winterhalder@uclouvain.be}\\
	\texttt{bellagente@thphys.uni-heidelberg.de}\\ \texttt{bpnachman@lbl.gov} \\
}

\begin{document}
	
	\maketitle

	\begin{abstract}
		Deep generative models are becoming widely used across science and industry for a variety of purposes. A common challenge is achieving a precise implicit or explicit representation of the data probability density.  Recent proposals have suggested using classifier weights to refine the learned density of deep generative models.  We extend this idea to all types of generative models and show how latent space refinement via iterated generative modeling can circumvent topological obstructions and improve precision. This methodology also applies to cases were the target model is non-differentiable and has many internal latent dimensions which must be marginalized over before refinement. We demonstrate our Latent Space Refinement (LaSeR) protocol on a variety of examples, focusing on the combinations of Normalizing Flows and Generative Adversarial Networks. We make all codes publicly available.\footnote{\url{https://github.com/ramonpeter/LaSeR}}
		%\footnote{\url{https://anonymous.4open.science/r/LaSeR}} % Anonymized for NeurIPS
	\end{abstract}
	
	\section{Introduction}
	
	Generative models are essential tools for many aspects of scientific and engineering workflows. First-principles simulations encode physical laws and then samples from these simulations can be used for designing, performing, and interpreting a measurement. However, these physics-based simulations can be too slow or not precise enough for a growing number of studies. Deep learning-based techniques such as Generative Adversarial Networks (GAN)~\cite{Goodfellow:2014:GAN:2969033.2969125,Creswell2018}, Variational Autoencoders (VAE)~\cite{kingma2014autoencoding,Kingma2019}, and Normalizing Flows (NF)~\cite{10.5555/3045118.3045281,Kobyzev2020} are powerful surrogates that can accelerate slow simulations and model complex datasets that would otherwise be intractable to describe from first principles. For example, a growing number of studies are exploring these tools for high energy physics (HEP) applications~\cite{deOliveira:2017pjk,Paganini:2017hrr,Paganini:2017dwg,Alonso-Monsalve:2018aqs,Butter:2019eyo,Martinez:2019jlu,Bellagente:2019uyp,Vallecorsa:2019ked,SHiP:2019gcl,Carrazza:2019cnt,Butter:2019cae,Lin:2019htn,DiSipio:2019imz,Hashemi:2019fkn,Chekalina:2018hxi,ATL-SOFT-PUB-2018-001,Zhou:2018ill,Carminati:2018khv,Vallecorsa:2018zco,Datta:2018mwd,Musella:2018rdi,Erdmann:2018kuh,Deja:2019vcv,Derkach:2019qfk,Erbin:2018csv,Erdmann:2018jxd,Urban:2018tqv,Oliveira:DLPS2017,deOliveira:2017rwa,Farrell:2019fsm,Hooberman:DLPS2017,Belayneh:2019vyx,buhmann2020getting,Alanazi:2020jod,2009.03796,2008.06545,Kansal:2020svm,Maevskiy:2020ank,Lai:2020byl,Choi:2021sku,Rehm:2021zow,Rehm:2021zoz,Carrazza:2021hny,Albergo:2019eim,Kanwar:2003.06413,Brehmer:2020vwc,Bothmann:2020ywa,Gao:2020zvv,Gao:2020vdv,Nachman:2020lpy,Choi:2020bnf,Lu:2020npg,Bieringer:2020tnw,Hollingsworth:2021sii,Monk:2018zsb,Cheng:2020dal,1816035,Howard:2021pos,Buhmann:2021lxj,Bortolato:2021zic,deja2020endtoend,Hariri:2021clz,Fanelli:2019qaq, 1800956, Bellagente:2021yyh}.
	
	A key difference in generative modeling between many scientific applications, including HEP, and typical industrial applications is that individual samples are often not useful. Inference is performed on a statistical basis and so it is essential to model the probability density precisely and not just match its support. Existing deep generative models have shown great promise in qualitatively modeling complex probability densities, but it is often challenging to achieve precision.
	
\begin{figure}[h!]
		\centering
		\includegraphics[width=0.95\textwidth]{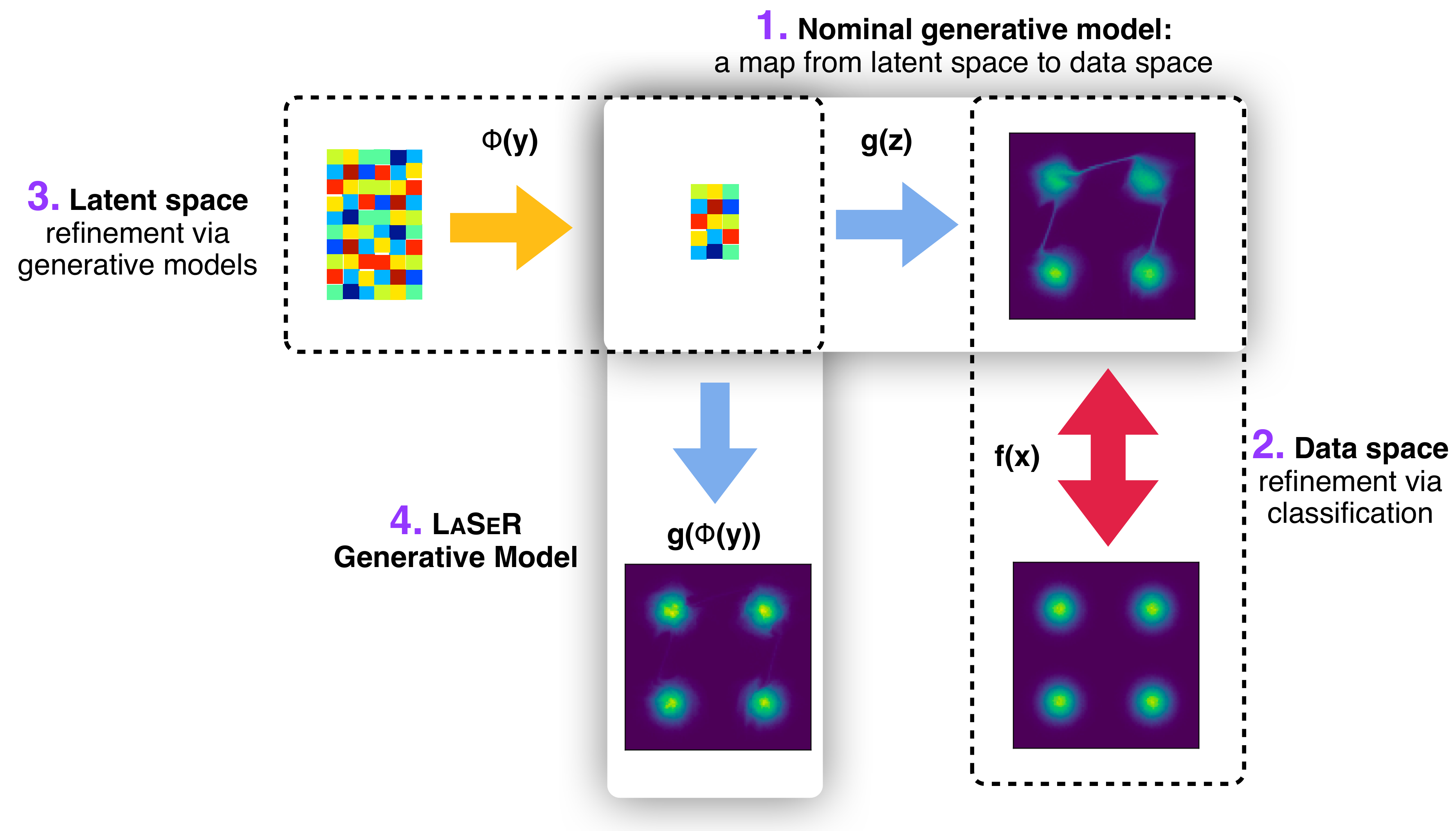}
		\caption{A schematic diagram illustrating the \textsc{LaSeR} protocol.  {\large\textbf{\color{Fuchsia}1}}. The upper right part of the diagram represents a given generative model $g:\mathbb{R}^N\rightarrow \mathbb{R}^M$ that maps features from a latent space to a data space.  {\large\textbf{\color{Fuchsia}2}}. A classifier network $f:\mathbb{R}^M\rightarrow \mathbb{R}$ is trained to distinguish between the generated samples and real samples from data.  The output of this classifier is interpreted as a likelihood ratio between the generated samples and the real samples and is pulled back to the latent space.  If $f$ is trained with binary cross entropy, the weights are approximated as $w(x)=f(x)/(1-f(x))$ and then the weight of a latent space point $z$ is $w(g(z))$.  {\large\textbf{\color{Fuchsia}3}}. Next, a second generative model $\Phi:\mathbb{R}^L\rightarrow \mathbb{R}^N$ is trained with these weights to transform the original latent space into a new latent space.  {\large\textbf{\color{Fuchsia}4}}. The \textsc{LaSeR} model is then given by $g(\Phi(y))$.}
		\label{fig:schematic}
	\end{figure}

	A useful strategy to improve the precision of generative models is to refine their predictions. For example, Ref.~\cite{che2020gan} showed how the classification part of a GAN can be used to reweight and resample the input probability distribution of the random noise. In Ref.~\cite{wu2020logan} this classifier information is already used during training to learn from an optimized latent space. A similar idea was introduced in Ref.~\cite{2009.03796} that is not specific to GANs, whereby a classifier network is trained on the generated samples to produce weights that improve the precision of the generated probability distribution. We combine and extent these approaches by introducing the \textbf{La}tent \textbf{S}pac\textbf{e} \textbf{R}efinement (\textsc{LaSeR}) protocol, illustrated in Fig.~\ref{fig:schematic}. Like \textsc{DctrGAN}~\cite{2009.03796,Andreassen:2019nnm}, \textsc{LaSeR} starts with any generative model $g(z)$ and trains a post-hoc classifier to distinguish the generated samples from the target data. A challenge with \textsc{DctrGAN} is that the results are weighted, which reduces the statistical power of a generated sample. The energy-based framework of Discriminator Driven Latent Sampling~(DDLS)~\cite{che2020gan} produces unweighted samples by transferring the weights to the latent space and then performing Langevin Markov Chain Monte Carlo (MCMC) to produce unweighted samples.  
	
	We develop a more general protocol that works for any generative model and is more efficient than the Langevin MCMC approach of Ref.~\cite{che2020gan}. We begin by pulling back the weights from our post-hoc classifier to the latent space, either directly or via the \textsc{OmniFold} method~\cite{Andreassen:2019cjw} when the generator is not surjective. We then propose to learn a second generative model $\Phi(y)$ that maps an auxiliary latent space onto the refined latent space. Generating from the refined model amounts to sampling from the auxiliary latent space and applying $g(\Phi(y))$. We focus on the case where $g$ is a normalizing flow and $\Phi$ is a GAN because GANs need not to be invertible
	%\footnote{There are tricks to achieve a similar result for Normalizing Flows~\cite{huang2020augmented}.}
	; however, the method can be applied to any pair of generative models. Furthermore, the procedure can be iterated for further refinement.  Learning a post-hoc generative model in a latent space was also studied by the authors of Refs.~\cite{bohm2020probabilistic,pmlr-v37-li15,xiao2019generative}, where a standard autoencoder becomes probabilistic via a second model such as a NF trained on the bottleneck layer. Similarly, the authors of Ref.~\cite{dai2018diagnosing} proposed an iterative VAE setup to bring the latent space closer to a target multidimensional Gaussian random variable. Further approaches in the context of VAEs include iterative refinement by Ref.~\cite{hjelm2018iterative}, iterative posterior inference by Refs.~\cite{marino2018iterative,kim2018semiamortized} as well as Ladder VAE~~\cite{sonderby2016ladder} which introduce a simple form of posterior revision.
	
	This paper is organized as follows. We review related work and describe the statistical properties and challenges associated with existing generative models in Section~\ref{sec:background}.  We introduce various ways of implementing the \textsc{LaSeR} protocol in Sec.~\ref{sec:methods}.  Illustrative numerical examples are provided in Sec.~\ref{sec:examples} and the paper ends with conclusions and outlook in Sec.~\ref{sec:conclusion}.
	
	%%%%%%%%%%%%%%%%%%%%%%%%%%%%%%%%%%%%%%%
	\section{Background}
	\label{sec:background}
	%%%%%%%%%%%%%%%%%%%%%%%%%%%%%%%%%%%%%%%
	A generator is a function $g$ that maps a latent space $\mathcal{Z}\subseteq \mathbb{R}^N$ onto a target or feature space $\mathcal{X}\subseteq \mathbb{R}^M$, with underlying probability densities $p_\mathcal{Z}$ and $p_\mathcal{X}$, respectively.  Typically, $p_\mathcal{Z}$ is chosen to be simple (e.g. normal or uniform) so that it is efficient to generate data $Z\sim p_\mathcal{Z}$.  In some cases, the latent space $\mathcal{Z}$ is broken into two components: one component that specifies the input features of interest and one component that specifies auxiliary latent features that are marginalized over.  When this happens, the map $g$ is viewed as a stochastic function of the first latent space component.  While our examples exclusively cover the case of deterministic maps from the full latent space to the target space, our approach can accommodate both settings as we explain in more detail below.
	
	The function $g$ can be constructed from first-principles insights about the dynamics of a system or it can be learned from data.  For example, commonly-used deep generative models include
	
	\begin{itemize}
		\item Generative adversarial networks (GANs)~\cite{Goodfellow:2014:GAN:2969033.2969125,Creswell2018}
		\item Variational autoencoders (VAEs)~\cite{kingma2014autoencoding,Kingma2019}
		\item Normalizing flows (NFs)~\cite{10.5555/3045118.3045281,Kobyzev2020} %and related  invertible neural networks (INNs)~\cite{ardizzone2019analyzing,1800956}.
		%\item Physics-based Monte-Carlo generators (e.g. from collider physics~\cite{Rambo, Platzer:2013esa, Alwall:2014hca, Sjostrand:2014zea, Frederix:2018nkq,Bellm:2015jjp, Bothmann:2019yzt}).
	\end{itemize}
	
	Apart from the specific learning objective, all generative models have their intrinsic advantages and disadvantages. The relevant features to understand the \textsc{LaSeR} protocol are summarized in Table~\ref{tab:comparison} and are further discussed in the following.
	
	\begin{table}[!htbp]
		\caption{A comparison of commonly used deep generative models.}
		\label{tab:comparison}
		\centering
		\begin{tabular}{lcccc}
			\toprule
			\parbox{3cm}{\centering Method} & \parbox{1.5cm}{\centering Train on data} & \parbox{1.5cm}{\centering Exact log-likelihood} & \parbox{2cm}{\centering  Non-topology preserving} \\ 
			\midrule
			Variational Autoencoders & {\color{green!70!black}\cmark} &  {\color{red!80!black}\xmark} & {\color{green!70!black}\cmark}\\
			Generative Adversarial Networks & {\color{green!70!black}\cmark} & {\color{red!80!black}\xmark} & {\color{green!70!black}\cmark}  \\
			Normalizing Flows & {\color{green!70!black}\cmark} & {\color{green!70!black}\cmark} & {\color{red!80!black}\xmark}\\
			\bottomrule
		\end{tabular}
	\end{table}
	
	%======================================
	\subsection{Generative models and coordinate transformations}
	\label{sec:event_generation}
	%======================================
	
	While the three generative models introduced in the previous section can all be trained directly on unlabeled data, they have different strategies for estimating $p_\mathcal{X}$.  GANs and VAEs learn this density implicitly by introducing an auxiliary task. In the case of GANs, the auxiliary task is performed by a discriminator network that tries to distinguish samples drawn from $p_\mathcal{Z}$ passed through $g$ and those drawn from $p_\mathcal{X}$ directly. For VAEs, the generator is called the decoder and the auxiliary task requires an encoder network $h$ to satisfy $g(h(x))\approx x$, while regularizing the latent space probability density.  Due to the structure of these networks, $N$ need not be the same size as $M$. 
	
	In contrast to GANs and VAEs, NFs explicitly encode an estimate for the probability density $p_\mathcal{X}$. These networks rely on a coordinate transformation which maps the prior distribution $p_\mathcal{Z}$ into a target distribution $p_g$ with $g$ now being invertible. This requires $M=N$ but allows for an analytic expression for the probability density induced by $g$:
	\begin{equation}
		p_{g}(x)\equiv p_g(g(z))=\left\vert\frac{\partial g(z)}{\partial z}\right\vert^{-1} p_\mathcal{Z}(z).
		\label{eq:coordinate_transform}
	\end{equation}
	In order to match $p_g$ and the data probability density $p_\mathcal{X}$, one can directly maximize the log-likelihood of the data without resorting to an auxiliary task:
	\begin{equation}
		\log p_{g}(x)=\log p_\mathcal{Z}(g^{-1}(x)) + \log \left\vert\frac{\partial g^{-1}(x)}{\partial x}\right\vert.
	\end{equation}
	
	%======================================
	\subsection{Topological obstructions}
	\label{sec:topology}
	%======================================
	
\begin{figure}[!htbp]
		\centering
		\includegraphics[width=0.4\textwidth]{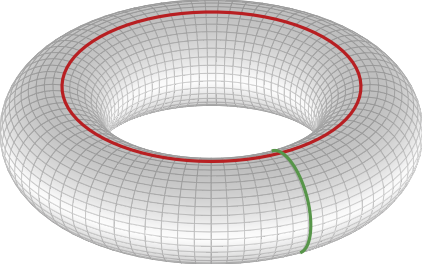}
		\caption{For a torus, the first Betti number is $b_1 = 2$ , which can be intuitively understood as the number of circular holes.}
		\label{fig:betti_torus}
	\end{figure}

	While the bijective nature of NFs allows for an explicit representation of the target probability density estimate, they inevitably suffer from a significant drawback.
	In order to find a mapping $g$ which satisfies Eq.~\eqref{eq:coordinate_transform} and matches the data probability density, the manifolds defined by the prior distribution $p_\mathcal{Z}$ and the target distribution $p_\mathcal{X}$ need to be topologically equivalent.
	A common way to describe the topological properties of manifolds are the Betti numbers.
	\begin{definition}
		The $n^\text{th}$ Betti number $b_n$ of a topological space $X$ is defined as the rank of the $n^\text{th}$ homology group $H_n$ of $X$~\cite{bettinumbers,hatcher2002algebraic}.
	\end{definition}
	Informally, the $n^\text{th}$ Betti number denotes the number of $n$-dimensional holes of a topological space.
	In fact, the first three Betti numbers do have simple geometrical interpretations:
	\begin{itemize}
		\item $b_0$ is the number of connected components.
		\item $b_1$ is the number of one-dimensional or circular holes.
		\item $b_2$ is the number of two-dimensional voids.
	\end{itemize}
	For instance, in Fig.~\ref{fig:betti_torus} we illustrate a torus which has one connected component $b_0 =1$, two one-dimensional holes $b_1 = 2$, and a single void enclosed within the surface $b_2 = 1$.
	%\medskip
	
	It has been proven by Poincar\'{e} that these numbers are invariant under homeomorphisms. This implies the following proposition:
	\begin{proposition}
		Any bijective mapping $g(z)$ is a homeomorphism and preserves the topological structure of the input space.
		\label{prop:topology}
	\end{proposition}
	A proof can be found in Refs.~\cite{younes2010shapes,dupont2019augmented}.
	This mathematical statement has a major impact on any generative model relying on bijective mappings, such as NFs. If the target space has a different topological structure than the latent space, these models cannot properly describe the target space topology as they inherently preserve the latent space topology.  It is indeed possible that these inconsistencies are hidden in some highly-correlated observable which might not be investigated during or after optimization. Moreover, it has been shown in Ref.~\cite{cornish2021relaxing} that you can hide and diminish the topological issues if you increase the network complexity. However, you can achieve better and computationally cheaper results by relaxing the bijetivity constraint itself~\cite{cornish2021relaxing} or by augmenting additional dimensions~\cite{dupont2019augmented, huang2020augmented}. Further, the issues arising by non-trivial topologies have also been investigated in the context of autoencoders in Ref.~\cite{Batson:2021agz}.
	
	Consequently, if we want to correctly reproduce the target space with a non-trivial topology, we either need to modify the latent space to match the topological structure, or the bijective model $g$ has to be replaced by a non-topology preserving model such as GANs or VAEs. While these models do have more flexibility and do not suffer from the same topological obstructions, this freedom usually goes hand in hand with training instabilities and more time-consuming hyperparameter searches. In order to benefit from both worlds and to fix the topology of the latent space, we propose to use the \textsc{LaSeR} protocol which is described in the following.
	
	%%%%%%%%%%%%%%%%%%%%%%%%%%%%%%%%%%%%%%%%
	\section{Proposed methods}
	\label{sec:methods}
	%%%%%%%%%%%%%%%%%%%%%%%%%%%%%%%%%%%%%%%%
	
	The \textsc{LaSeR} protocol is illustrated in Fig.~\ref{fig:schematic}. The input is a generator $g(z)$ that could either be a neural network or a black-box (BB) simulator (e.g. a physics-based Monte Carlo). 
	
	As a first step, a classifier is trained to distinguish samples drawn from $g$ and samples drawn from the target data probability density. It is well-known~\cite{hastie01statisticallearning,sugiyama_suzuki_kanamori_2012} (and forms the basis of the \textsc{Dctr} protocol~\cite{1907.08209}) that an ideally trained classifier $f$ using the binary cross entropy (BCE) loss function will have the property $f/(1-f)\propto p(x|\text{target})/p(x|\text{from $g$})$. The proportionality constant is $p(\text{target})/p(\text{from $g$})$ and is unity when the two datasets have the same number of samples. Other loss functions such as the mean squared error have the same property. It is also possible to engineer the loss function to directly learn the likelihood ratio~\cite{DAgnolo:2018cun,Nachman:2021yvi}. Throughout this paper, we use BCE.
	
	Each generated sample is assigned a weight $w(x_i)=f(x_i)/(1-f(x_i))$. These weights are then assigned to the corresponding latent space point $g(z_i)=x_i$. When $g$ is surjective, $w$ is a proper function of $z$. However, there are many potential applications where the generator has many internal latent degrees of freedom that are marginalized over and are not part of $\mathcal{Z}$. In this case, one can use the \textsc{OmniFold} algorithm~\cite{Andreassen:2019cjw} which is an iterative expectation maximization-style approach that returns weights for the $z_i$ that maximizes the likelihood in the data space.  The weights extracted with \textsc{OmniFold} are a proper function of the latent space. This is not necessarily the case for weights directly inferred from the classifier, since the mapping $g$ can send the same latent space point to different data space points. In other words, the marginalized latent dimensions make $g$ a stochastic function.

	The final step of \textsc{LaSeR} requires to sample from weighted latent space $(z,w(z))$. This can be for instance done directly using the following methods:
	
	\begin{itemize}
		\item \textbf{Rejection sampling:}
		We can use the weights $w(z)$ to perform rejection sampling on the latent space. This is simple, but ineffective if the weights are broadly distributed. Thus, we do not employ this method in our experiments.
		\item \textbf{Markov chain Monte Carlo (MCMC):} 
		The weights $w(z)$ induce a new probability distribution
		\begin{equation}
			q_\mathcal{Z}(z)= p_\mathcal{Z}(z)\,w(z)/Z_0,
			\label{eq:mcmc_distribution}
		\end{equation}
		with some normalisation factor $Z_0$. In general, this probability distribution is non-trivial and does not allow to determine its quantile function analytically. 
		However, if the probability distribution $q_\mathcal{Z}$ and its derivative $\partial q_\mathcal{Z}/\partial z$ are tractable a MCMC algorithm can be employed. While Ref.~\cite{che2020gan} uses a simple Langevin dynamics algorithm, we suggest to use the more advanced Hamiltonian Monte Carlo (HMC)~\cite{DUANE1987216,neal2012mcmc} algorithm as it generates significantly better samples.
	\end{itemize}
	
	However, instead of using the above approaches, we suggest for the \textsc{LaSeR} protocol to convert the weighted latent space $(z,w(z))$ into a new unweighted latent space.
	For this, we train a second generative model $\Phi$ which we call the refiner. The refiner maps an auxiliary latent space $\mathcal{Y}\subseteq\mathbb{R}^L$ onto the refined latent space $\mathcal{Z}^\prime\subseteq\mathbb{R}^N$, with underlying probability densities $p_\mathcal{Y}$ and
	$p_{\mathcal{Z}^\prime}=q_\mathcal{Z}$, respectively.
	The refiner $\Phi$ can be either a GAN or a NF, where the latter requires $L=N$. In order to train the model on the weighted latent space $(z,w(z))$ we accommodate the weights into the loss functions as it was proposed in Refs.~\cite{Backes:2020vka, Verheyen:2020bjw}. In contrast to the MCMC algorithm, this method does not require to calculate the derivative of the weights $\partial w/\partial z$.
	
	At the end, we generate new samples from the refined generator by first sampling from the auxiliary latent space, mapping that to the refined latent space, and then passing that through $g$. In some cases, the entire procedure can be iterated, i.e.\,if $g$ is a neural network, its weights could be updated using the refined latent space.
	%%%% RW: until here %%%%%
	
	Table~\ref{tab:refiner_comparison} illustrates some of the features of various combinations of generators $g$ and latent space refiner networks $\Phi$.
	
	\begin{table}[!htbp]
		\caption{Comparison of various combinations of generator $g$ and refiner network $\Phi$.}
		\label{tab:refiner_comparison}
		\centering
		\begin{tabular}{ccl}
			\toprule
			\parbox{2cm}{\centering Generator $g$} & \parbox{2cm}{\centering Refiner $\Phi$} & \parbox{5cm}{Note}\\ 
			\midrule
			GAN & GAN &  Possible to iterate \\
			GAN & NF &  Possible to iterate \\
			NF & GAN &  Cannot iterate (density no longer explicit)\\
			NF & NF &  Suffers topological obstructions \\
			\bottomrule
		\end{tabular}
	\end{table}

	% \textbf{Should cite Ref.~\cite{bohm2020probabilistic} somewhere.  This is a little similar to what we do, although totally different.} -> add to introduction

	%Describe the laser protocoll with its possible variants(
	%Primary Generator + Refiner):
	%\begin{itemize}
	%    \item GAN + GAN
	%    \item GAN + NF (doesn't fix topology if not already fixed in GAN)
	%    \item NF + GAN (can't retrain if using MLE)
	%    \item Physics Generator(MC) + GAN/NF
	%    \item GAN/NF/MC + Markov Chain
	%    \item GAN/NF/MC + Rejection Sampling (in principle)
	%\end{itemize}
	
	%Give some short recap of how the weights are extracted with a classifier (DCTR-approach, and Discriminator driven). These weights can be either directly pulled back onto the latent space of if the generator is not surjective using OmniFold. Then we can
	%use this weighted latent space and gain an unweighted latent space in three possible ways:
	%\begin{itemize}
	%    \item Use the weights to do rejection sampling on the latent space. Ineffective if the weights are broadly distributed.
	%    \item Use Markov Chain Monte Carlo - Langevin Dynamics < MALA (Metropolis-adjusted Langevin algorithm) < HMC (Hamiltonian Monte Carlo)
	%    \item Train a second generative model (GAN or NF)
	%    to learn the unweighted space from the weighted one (cite uwGAN paper and the weightedFlow paper) 
	%    [ Thoughts: in principle, this might also work with a VAE? Not sure]
	%\end{itemize}

	%%%%%%%%%%%%%%%%%%%%%%%%%%%%%%%%%%%%%%%%
	\section{Numerical examples}
	\label{sec:examples}
	%%%%%%%%%%%%%%%%%%%%%%%%%%%%%%%%%%%%%%%%
	
	We consider three different 2-dimensional examples which allows to visualize both the full output and the latent space of the primary generator.
	All three sets of training data consist of 480k data points. In all models and experimentes we used a batch size of 2000 points and an epoch is defined as iterating through all points in the training set, giving 240 model updates per epoch. We run all our experiments on
	our internal GPU cluster, which consists of multiple Nvidia GTX 1080 TI and Nvidia RTX 2080 Ti GPUs. Depending on the model the computation times range between 30 mins and 5 hours.
	
	\subsection{Implementation details}
	\label{sec:implementation_details}
	
	All models and the training routines are implemented in \textsc{PyTorch} 1.8~\cite{Pytorch2019}. In the following we specify the implementation details for the various networks and the chosen hyperparameters.
	
	\paragraph{Baseline model.}
	In all examples we use the Real-NVP~\cite{RNVP2019} implementation of a NF as the baseline model. Our implementation relies partially on the \textsc{FrEIA} library.\footnote{Framework for Easily Invertible Architectures: \texttt{https://github.com/VLL-HD/FrEIA}}
	In the first two examples, the NF consists of 12 coupling blocks, where each block consists of a fully connected multi-layer perceptron (MLP) with 3 hidden layers, 48 units per layer and leaky ReLU activation functions. In the third more complex example, we increase the number of coupling blocks to 20 and the number of units to 60. In all cases the model is trained for 100 epochs by minimizing the negative log-likelihood with an additional weight decay of $10^{-5}$. The optimization is performed using Adam~\cite{Adam2019} with default $\beta$ parameters and $\alpha_0 = 10^{-3}$. We further employ a exponential learning rate schedule, i.e.\,$\alpha_n = \alpha_0\,\gamma^n$,
	with $\gamma = 0.999$ which decays the learning rate after each epoch.
	
	\paragraph{Post-hoc classifier.}
	As a post-hoc classifier we employ a MLP with 8 hidden layers, 96 units per layer and leaky ReLU activation functions. We train it for 50 epochs by minimizing the BCE loss between true data samples and samples from the generator $g$. Optimization is again performed using Adam with default $\beta$ parameters and $\alpha_0 = 10^{-3}$ and an exponential learning rate schedule with $\gamma = 0.999$.
	
	\paragraph{Refiner network.}
	Finally, the refiner network is a GAN consisting of a discriminator and a generator, both MLPs with 7 hidden layers, 100 units per layer and Leaky ReLU activation functions.
	As a GAN does not require to have the same dimensions in the latent space as in the feature space we choose 4-dimensional auxiliary latent space to simplify the task of the generator. The standard GAN objective is replaced by the weighted BCE loss introduced in Ref.~\cite{Backes:2020vka} to accommodate the weights of the weighted training data. The GAN is trained for 200 epochs using the Adam optimizer with $(\alpha, \beta_1, \beta_2) = (10^{-4}, 0.5, 0.9)$ and also using an exponential learning rate schedule with $\gamma = 0.999$. To compensate an imbalance in the training we update the discriminator four times for each generator update.
	
	\paragraph{Hamiltonian Monte Carlo (HMC).}
	In order to have full control of all parameters and inputs we implemented our own version of Hamiltonian Monte Carlo in \textsc{PyTorch}, using its automatic differentiation module to compute the necessary gradients to run the algorithm.
	For all examples, we initialize a total of 100 Markov Chains running in parallel to reduce computational overhead and solve the equation of motions numerically using the leapfrog algorithm~\cite{neal2012mcmc} with a step size of $\varepsilon=0.004$ and $50$ leapfrog steps.
	To avoid artifacts originating from the random initialization of the chains, we start with a burn-in phase and discard the first 3000 points from each chain.
	
	\subsection{Probability distance measures}
	\label{sec:distance_measure}
	
	In order to quantify the performance of the base model and its refined counterparts, we implemented two measures of distance:
	
	\paragraph{Jensen--Shannon divergence (JSD).}
	
	A simple but commonly used distance measure between two probability distributions $p(x)$ and $q(x)$ is the Jensen–Shannon divergence
	\begin{equation}
		\mathrm{JSD}(p,q) =  \frac{1}{2}\int dx\,p(x)\log\left(\frac{2p(x)}{p(x)+q(x)}\right)+q(x)\log\left(\frac{2p(x)}{p(x)+q(x)}\right).
	\end{equation}
	In contrast to the Kullback–Leibler (KL) divergence it is symmetric and always returns a finite value.
	
	\paragraph{Earth mover distance (EMD).}
	
	Another common distance measure between two probability distributions $p(x)$ and $q(x)$ is the earth mover's distance (EMD). The EMD is usually formulated as an optimal transport problem. Lets assume the two distributions are represented as clusters $P=\{(x_i,p(x_i)\}_{i=1}^M$ and $Q=\{(y_j,q(y_j)\}_{j=1}^{M^\prime}$. Then, the EMD between two probabilities is the minimum cost required to rearrange one probability distribution $p$ into the other $q$ by discrete probability flows $f_{ij}$ from one cluster point $x_i$ to the other $y_j$
	\begin{equation}
		\begin{split}
			\mathrm{EMD}(p,q)&=\min_{\{f_{ij}>0\}}\sum_{i=1}^M \sum_{i=1}^{M^\prime}\,f_{ij}||x_i - y_j||_2, \\
			\sum_{j=1}^{M^\prime}\,f_{ij}\leq p(x_i),\quad \sum_{i=1}^{M}&\,f_{ij}\leq q(y_j),\quad
			\sum_{i=1}^M \sum_{i=1}^{M^\prime}\,f_{ij}=\min\left(\sum_{i=1}^{M}p(x_i), \sum_{j=1}^{M^\prime}q(y_j)\right),
		\end{split}
	\end{equation}
	where the optimal flow $f_{ij}$ is found by solving this optimization problem. For an efficient implementation we used the \textsc{POT} library~\cite{flamary2021pot}.
	
	\paragraph{Numerical approximation.}
	In order to calculate these quantities numerically we generate 2M data points for all models and examples and approximate the probability densities from a 2-dimensional histogram. As ground distance measure between clusters $P$ and $Q$ for the calculation of the EMD, we consider the Euclidean distance between the center points of each bin. In order to estimate the error by approximating the probabilities, we calculated the JSD and EMD score on equally large but independent test samples drawn from the truth distribution. The extracted error estimations are indicated in parenthesis in Table~\ref{tab:scores}.
	
	\subsection{Experimental results}
	\label{sec:results}
	
	In the following we will simply refer to the \textsc{LaSeR} protocol using the Hamiltonian Monte Carlo as HMC method, and we will refer to the \textsc{LaSeR} protocol using a refiner network as \textsc{LaSeR} method.
	
\begin{figure}[!htbp]
		\centering
		\includegraphics[width=0.85\textwidth]{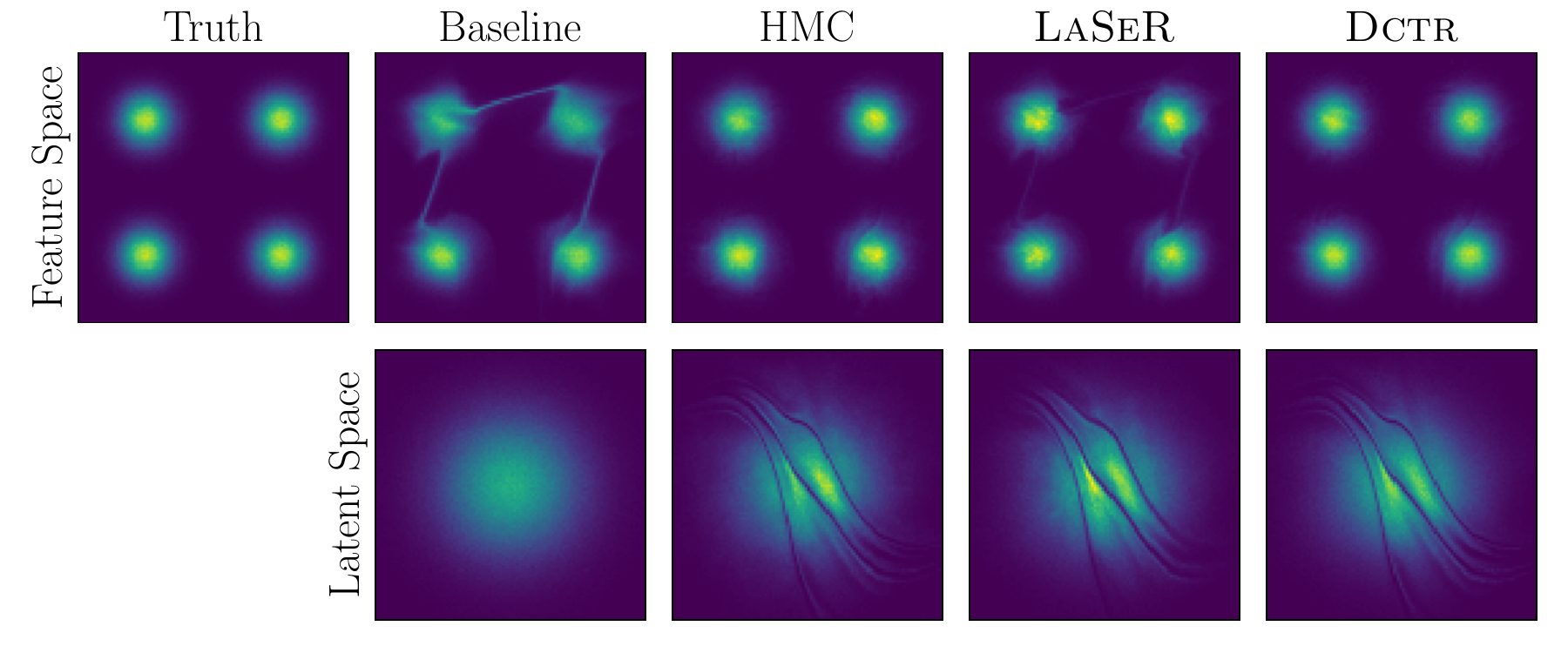}
		\caption{Comparison of the baseline model and the refined outputs by either using the HMC, \textsc{LaSeR} or \textsc{Dctr} method on a 2-dimensional multi-modal gaussian distribution with Betti numbers $b_0=4$, $b_1=0$.}
		\label{fig:gaussians}
	\end{figure}

	In the first example, illustrated in Fig.~\ref{fig:gaussians}, we considered a multi-modal distribution which can be topologically described by its Betti numbers $b_0=4$ and $b_{n\ge1}=0$. In contrast, our non-refined latent space is described by a 2-dimensional normal distribution with $b_0=1$ and $b_{n>0}=0$. As both spaces are topologically different our baseline model fails to correctly reproduce the truth distribution. It is obvious that the baseline model cannot properly separate the individual modes and always leaves a connection between them. This is in agreement with Proposition \ref{prop:topology} as our baseline model is a NF and thus inevitably topology preserving.
	After training a post-hoc classifier the reweighted feature space of the \textsc{Dctr} method clearly resolved the topological obstructions. Pulling back these weights onto the orginal latent space induces a weighted latent space which is now also clearly separated into four pieces and thus shows the same topological structure as the data distribution. This weighted latent space is correctly reproduced in the refined latent spaces of the HMC and \textsc{LaSeR} method. Acting with our generator $g$ on these refined latent spaces shows a highly improved feature space distribution.

\begin{figure}[!htbp]
		\centering
		\includegraphics[width=0.85\textwidth]{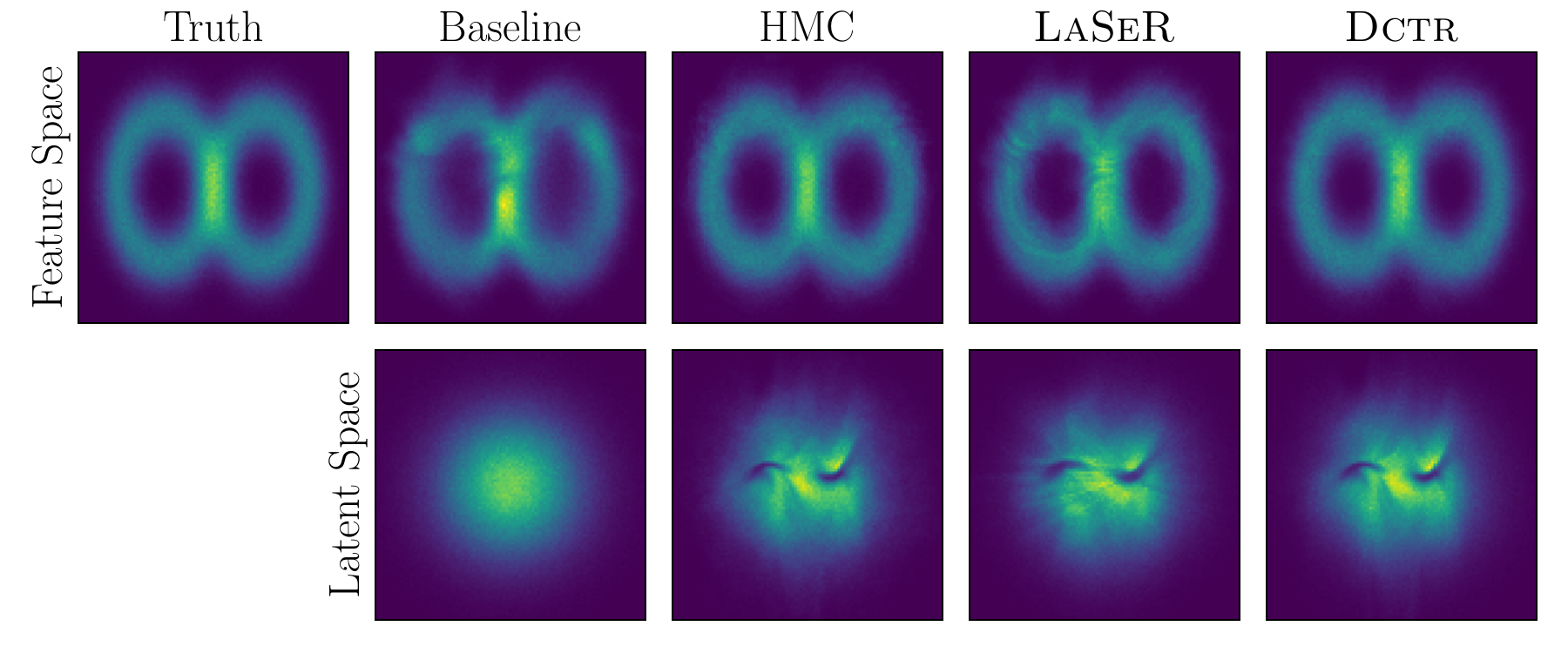}
		\caption{Comparison of the baseline generator and the refined outputs by either using the HMC, \textsc{LaSeR} or \textsc{Dctr} method on a 2-dimensional double donut distribution with Betti numbers $b_0=1$, $b_1=2$.}
		\label{fig:donut}
	\end{figure}

	In Fig.~\ref{fig:donut}, we considered a double donut distribution with Betti numbers $b_0=1$, $b_{1}=2$ and $b_{n\ge2}=0$ as a second example.
	This example does not have disconnected parts but contains two circular holes. Again the baseline model has troubles to properly close the loops and reproduce the holes as these topological features are not present in the non-refined latent space. After reweighting the latent space with the \textsc{Dctr} weights, these topological features emerge as holes in the latent space. Both refined latent spaces show the same topological structure as the reweighted latent space, resulting in an improved feature space distribution for both the HMC and the \textsc{LaSeR} method.

\begin{figure}[!htbp]
		\centering
		\includegraphics[width=0.85\textwidth]{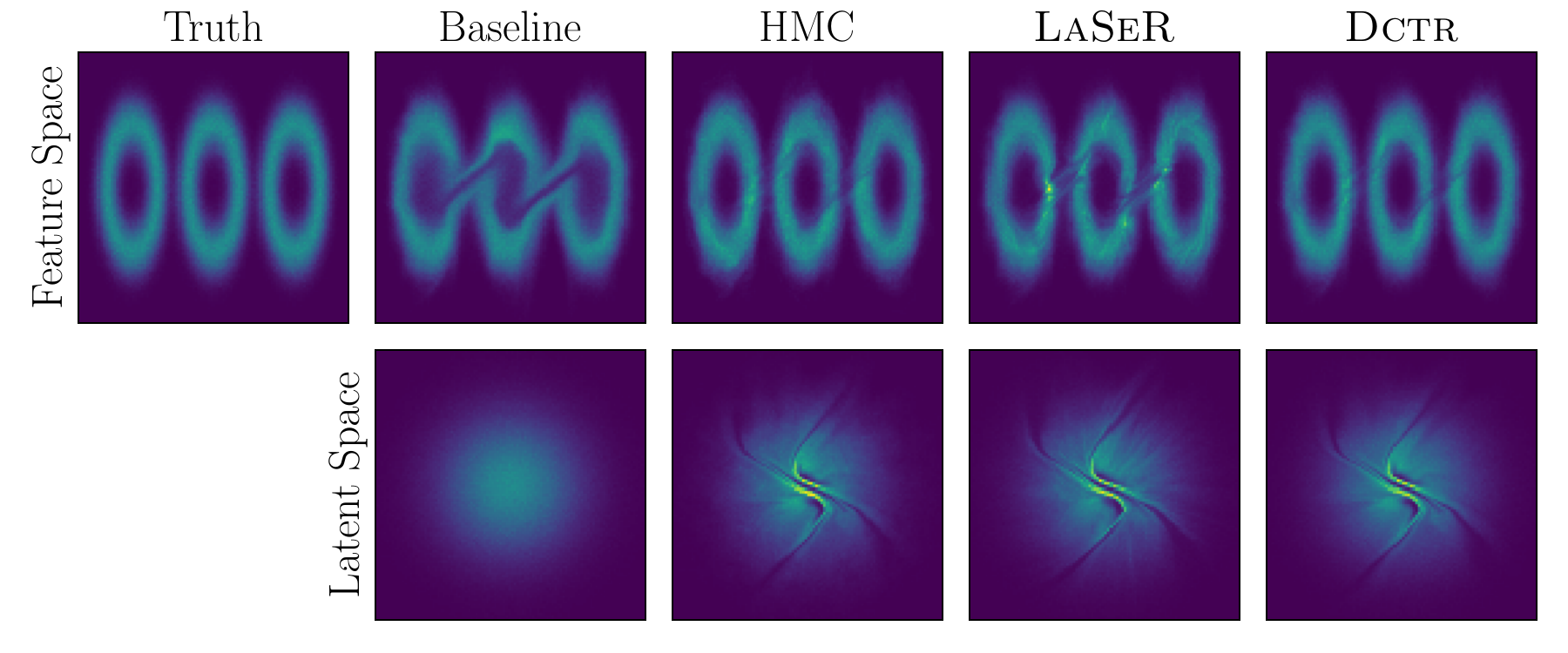}
		\caption{Comparison of the baseline generator and the refined outputs by either using the HMC, \textsc{LaSeR} or \textsc{Dctr} method on a 2-dimensional triple ring distribution with Betti numbers $b_0=3$, $b_1=3$.}
		\label{fig:3_rings}
	\end{figure}

	Finally, in the last example we combine the topological features from the previous examples to make our baseline model fail even more. In detail, we considered a set of 3 displaced rings which can be topologically described by the Betti numbers $b_0=3$, $b_{1}=3$ and $b_{n\ge1}=0$, as illustrated in Fig.~\ref{fig:3_rings}. As with previous examples, the baseline model utterly fails to reproduce the true data distribution. As the topological structure of the data distribution is more complex the reweighted latent space in the \textsc{Dctr} method is considerably more involved then the weighted latent spaces of the previous examples. Owing to this more complex structure of the weighted latent space our refiner model $\Phi$ has a much harder job to properly reproduce these topological features in the refined latent space of the \textsc{LaSeR} method. While being far from optimal the feature space distribution of the \textsc{LaSeR} refined generator is notably improved in comparison to the baseline model. In this particular complicated example the output of the HMC method seems to show slightly better agreement with the truth distribution than the \textsc{LaSeR} method. However, these results can be improved by spending more time on a detailed network hyperparameter optimization.
	\begin{table}[!htbp]
		\caption{Earth mover distance (EMD) and Jensen–Shannon divergence (JSD) on test data for the baseline model and the proposed refinement methods for various two-dimensional examples. The best results are written in bold face. The errors on the approximate EMD and JSD are indicated in parenthesis.}
		\label{tab:scores}
		\centering
		\begin{tabular}{lcccccc}
			\toprule
			\multirow{2}*{Method}& \multicolumn{2}{c}{Gaussians} & \multicolumn{2}{c}{Double donut} & \multicolumn{2}{c}{Rings}\\
			\cmidrule(r){2-3} \cmidrule(r){4-5} \cmidrule(r){6-7}
			& EMD  & JSD & EMD & JSD & EMD & JSD\\
			\midrule
			Baseline & $0.28(2)$  & $0.66(4)$  & $0.23(2)$  & $0.27(4)$ & $0.067(6)$  & $0.21(2)$  \\
			HMC & $0.22(2)$ & $\mathbf{0.12(4)}$ & $\mathbf{0.06(2)}$  & $\mathbf{0.07(4)}$ & $0.137(6)$  & $\mathbf{0.07(2)}$ \\
			\textsc{LaSeR} & $\mathbf{0.21(2)}$ & $0.24(4)$ & $0.09(2)$  & $0.09(4)$ & $\mathbf{0.047(6)}$  & $0.08(2)$ \\
			\midrule[0.2pt]
			\textsc{Dctr}  & $0.09(2)$ & $0.09(4)$ & $0.11(2)$  & $0.05(4)$ & $0.037(6)$  & $0.04(2)$ \\
			\bottomrule
		\end{tabular}
	\end{table}

	In order to provide a quantitative comparison of performances, we calculated the EMD and JSC scores of our baseline model and its refined counterparts for all examples. These scores are summarized in Table~\ref{tab:scores} and show that the \textsc{HMC} method yields the best performance in most scenarios for the unweighted output. However, the \textsc{LaSeR} method is equally good or better for the the most complex example, showing that the HMC method has troubles to sample from the highly tangled latent space.  Moreover, the HMC method cannot be used at all if the derivatives of the weights are not tractable, in which case the \textsc{LaSeR} method is still applicaple. The overall best performance is achieved by the \textsc{Dctr} method which however corresponds to a weighted feature space. As these weights are also used to obtain the refined latent space for the HMC and \textsc{LaSeR} method the score of the \textsc{Dctr} yield as a benchmark and provides lower limit.
	Indeed, we can ask ourselves why the EMD of both unweighted methods are smaller than the EMD of the \textsc{Dctr} method for the double donut example. However, owing to the
	the uncertainty on the numerical estimation of the scores, which are indicated in parenthesis, all scores a statistically consistent.

	%%%%%%%%%%%%%%%%%%%%%%%%%%%%%%%%%%%%%%%
	\section{Conclusions and outlook}
	\label{sec:conclusion}
	%%%%%%%%%%%%%%%%%%%%%%%%%%%%%%%%%%%%%%%
	
	We have presented the \textsc{LaSeR} protocol, which provides a post-hoc method to improve the performance of any generative model. In particular, it has shown remarkable results on several numerical examples and improved the EMD (JSD) score of the baseline model by 25-61\% (62-67\%). It further managed to fix the topological obstructions encountered by the base model.  While we have only explicitly demonstrated \textsc{LaSeR} for refining normalizing flows as the primary generator, the protocol is also valid for other deep generative models as well as other generative models that need not be differentiable or surjective.
	
	While \textsc{LaSeR} is a promising approach for generative model refinement, it also has some limitations.  First of all, there is some dependence on the efficacy of the primary generator. For example, if the primary generator is a constant, then no amount of refining can recover the data probability density.  Furthermore, the refinement comes at a computational cost with an additional step on top what is required to construct the original generative model.  This can be particularly important when the new latent space is sourced from a much bigger space than the original one.  Finally, even though \textsc{LaSeR} produces unweighted examples, the statistical precision of the resulting dataset may be limited by the size of the training dataset~\cite{2008.06545}.  We leave further investigations of the statistical power of refined samples and the interplay with hyperparameter choices for future research.
	
	%%%%%%%%%%%%%%%%%%%%%%%%%%%%%%%%%%%%%%%
	\begin{ack}
		BPN was supported by the Department of Energy, Office of Science under contract number DE-AC02-05CH11231. RW acknowledges support by HeiKA and by FRS-FNRS (Belgian National Scientific Research Fund) IISN projects 4.4503.16.
		MB is supported by the International Max Planck School Precision Tests of Fundamental Symmetries.  We thank Tilman Plehn, Uro\v{s} Seljack, and Jesse Thaler for their helpful feedback on the manuscript.
	\end{ack}
	
	%%%%%%%%%%%%%%%%%%%%%%%%%%%%%%%%%%%%%%%
	\bibliographystyle{JHEP.bst}
	\bibliography{main,HEPML}
	
	\clearpage

	\appendix
	
\end{document}